%% file: main.tex
\documentclass[conference]{IEEEtran}
\usepackage{graphicx}
\usepackage{amsmath}
\usepackage{amssymb}
\usepackage{mathtools}
\usepackage{bm}
\usepackage{comment}
\usepackage{subfig}
\usepackage{wrapfig}
\usepackage{booktabs} 

\usepackage[font=small,labelfont=bf]{caption}

\usepackage{array}
\newcolumntype{P}[1]{>{\centering\arraybackslash}p{#1}}
\newcolumntype{M}[1]{>{\centering\arraybackslash}m{#1}}

\def\rrr#1\\{\par
\medskip\hbox{\vbox{\parindent=2em\hsize=6.12in
\hangindent=4em\hangafter=1#1}}}

\begin{document}

\input{frontmatter-sv}

\begin{abstract}
\input{abstract}
\end{abstract}

\section{Introduction}
\label{sec:introduction}
\input{introduction}

\section{Related Work}
\label{sec:relatedwork}
\input{related}

\section{Network Multivariate Time Series Forecasting}
\label{sec:problem}
\input{problem}

\section{Deep Learning Models for Multi-Variant Time Series Forecasting}
\label{sec:solution}
\input{solution}

\section{Experiments and Evaluation}
\label{sec:evaluation}
\input{evaluation}

\section{Conclusions}
\label{sec:future}
\input{future}

\bibliographystyle{IEEEtran}
\bibliography{tmgpt,network}


\end{document}

%% file: frontmatter-sv.tex
\title{Network-Temporal Learning for Large-Scale Network Traffic Prediction}
\thanks{This work was supported by the
    US National Science Foundation under Grant OAC-183990.
}
\author{\IEEEauthorblockN{Yufeng Xin, Ethan Fan}
\IEEEauthorblockA{
RENCI, University of North Carolina at Chapel Hill\\
Chapel Hill, NC, USA\\
}
}

\maketitle
\thispagestyle{empty}

%% file: abstract.tex
Accurate prediction of multivariate time series is essential for emerging network intelligent control, observability, and management functions. Existing statistical-based and shallow machine learning models have shown limited prediction capabilities on multivariate time series. They prioritize improvements in average prediction accuracy, while overlooking heterogeneous dependency structures and performance variability across individual time series. Recent advances in large language models have introduced new directions for multivariate time series forecasting; however, their application in conjunction with explicit structural dependency modeling remains relatively underexplored, especially in networked environments.

In this paper, we present a topology-aware learning framework for large-scale network traffic prediction that explicitly models both temporal dynamics and structural dependencies in multivariate network time series. We first investigate a graph attention model designed to capture topology-induced correlations among network traffic time series. We then evaluate a fine-tuned large language model-based representations for improved generalization across heterogeneous traffic patterns. To further address the diversity of cross-correlations in high-dimensional traffic data, we introduce a clustering-based preprocessing stage that groups traffic flows with similar dependency characteristics prior to model training, reducing input complexity and improving learning stability.
Experiments on real backbone traffic data show consistent improvements over statistical and recurrent neural network baselines. In addition to average accuracy, we evaluate performance across individual time series and observe reduced variability in prediction quality.

%% file: introduction.tex
Accurate prediction of network traffic dynamics is essential for traffic engineering, capacity planning, and anomaly detection, prediction, and mitigation in large-scale networked systems~\cite{plan:drl:zhu2021,ilbert2023breaking,singh2022traffic,shaghaghi2021proactive,tanaka2021monitoring}. These problems are naturally formulated as multivariate time series forecasting tasks, where multiple traffic streams exhibit complex and time-varying dependencies. While a range of statistical and deep learning models have been proposed, most existing approaches emphasize improvements in average prediction accuracy and provide limited insight into performance variability across individual time series. This limitation becomes more pronounced in high-dimensional settings, where correlations among traffic flows are heterogeneous and evolve over time, often leading to uneven prediction quality across the network. Recent advances in large language models have introduced new capabilities for sequence modeling, but their use in conjunction with explicit modeling of inter-series dependencies remains limited in network traffic prediction.

To address these challenges, we develop a topology-aware temporal learning framework that explicitly incorporates both structural dependencies and temporal dynamics in multivariate network traffic data. The proposed approach includes a network-temporal graph attention model for capturing topology-induced correlations and a a fine-
tuned large language model for improved accuracy and generalization. To handle heterogeneous dependency patterns across traffic flows, we introduce a clustering-based preprocessing step that groups time series with similar correlation structures prior to model training. This design improves scalability and stabilizes learning in high-dimensional settings. Furthermore, we evaluate prediction performance not only in terms of average error but also across the distribution of individual time series, providing a more comprehensive assessment of model behavior in large-scale networks.

Traffic measurements obtained on different parts of a 
network naturally form a multivariate time series (MTS) with 
measurements from multiple network devices. Network traffic data often entails complex statistical and temporal patterns, such as non-stationary and non-normality, and a mixture of long or short-term seasonality~\cite{lai2018modeling}. 
The fact that these time series are interconnected in a network setting suggests that the spatial dependencies can be potentially leveraged to improve prediction performance in addition to the temporal pattern recognition.  
The conventional statistical models require pre-defined global parameters that are both hard to learn and to generalize to non-stationary time series with complex temporal patterns. As such, deep neural networks (DNNs) have been dominating research in multivariate time-series analysis, particularly in spatial-temporal models. The stunning success of {\it large language models} (LLMs), {\it aka} {\it pre-trained foundation models} (FMs), in their originally intended natural language processing (NLP) and computer vision (CV) applications, has drawn immediate attention from time series researchers due to their inherent strength in dealing with sequential data.

One major limitation is that most time series benchmark datasets feature small numerical values within narrow ranges, simple seasonality patterns, and a low number of time series, which can not match the scale and capture the topological structure of large-scale network data. Aside from the common explainability complaints associated with NN models, training complexity and generalizability of these models remain great challenges. Their applicability in network prediction tasks has not been studied. Furthermore, several common pitfalls in selecting adequate performance metrics, cross-validation, data scaling, prediction horizon, and information leakage have been largely ignored, which would deem their conclusions and claimed performance superiority unreliable ~\cite{ts:forecast:pitfalls:2023}. Existing research in deep-learning based time series analysis has not targeted the specific intricate topological interdependency that network time series data in particular presents. 

In this paper, we investigate topology-informed temporal learning architectures that capture both spatial dependency and temporal dynamics in large-scale network traffic data, enabling improved prediction performance in heterogeneous network environments. We introduce a clustering-based preprocessing approach that groups traffic flows with similar dependency characteristics prior to model training, reducing input complexity and improving model stability for high-dimensional network traffic prediction. In addition to conventional average error metrics, we analyze the distribution of prediction accuracy across individual time series to characterize model behavior under heterogeneous traffic conditions, providing a more comprehensive assessment of prediction reliability. We conduct evaluation across multiple prediction horizons and compare the proposed approach with classical statistical models and recurrent neural network baselines, demonstrating consistent improvements in both average accuracy and performance stability.
Experimental results demonstrate consistent improvements over classical statistical models and recurrent neural network baselines in average and distribution-level prediction performance over different horizons. These findings highlight the importance of incorporating topology-aware modeling, dependency-aware preprocessing, and distribution-sensitive evaluation for scalable and reliable traffic prediction in next-generation networks.

The rest of this paper is organized as follows. We first present related work on multivariate time series prediction models in Section~\ref{sec:relatedwork}. We then describe the network-temporal time series forecasting problem in Section~\ref{sec:problem}. 
In Section~\ref{sec:solution}, we present a network-temporal graph attention network (NT-GAT) model and a fine-tuned cross-modal LLM model enhanced with an MTS clustering pre-training step. 
The performance evaluation via extensive numerical analysis on a real-world network traffic trace
is presented in Section~\ref{sec:evaluation}. The paper is concluded in Section~\ref{sec:future}.

%% file: related.tex
Classical statistical time series prediction models can be categorized into time-domain and frequency-domain methods. The SARIMAX (Seasonal Autoregressive Integrated Moving Average + exogenous variables) model is conceivably a suitable time-domain prediction model with both trends and seasonal variations while incorporating exogenous variables into the analysis to improve prediction accuracy. 

Popular DNN models such as Multilayer Perceptron (MLP), Long Short Term Memory (LSTM), convolution neural network (CNN), and graph neural network (GNN) have shown improved prediction performance in certain multivariate and spatial-temporal time-series applications with their increased modeling capability~\cite{lai2018modeling,ali2024resource}. The spatial-temporal graph attention network (ST-GAT) stacks a geometry distance-based GAT model and an LSTM model together to capture the spatial dependency and temporal pattern in sequence. The ST-GAT model has shown state-of-the-art (SOTA) performance in transportation system applications~\cite{st_gan}.
 
Evolving from the univariate time series-based models, a few recent studies have attempted multivariate and spatial-temporal approaches with classical statistical methods and simple neural networks in predicting wide area network (WAN) traffic volumes~\cite{mohammed2021predicting} and wireless network traffic~\cite{ilbert2023breaking}. 

LLMs' capabilities to represent complex patterns in sequence data and enable zero-shot or few-shot learning have catalyzed a new wave of application development in time series classification and forecasting~\cite{garza2023timegpt,NEURIPS2023_3eb7ca52,jin2024time}, often building upon work done in applying transformers to time series analysis. The major research interest has been the model's capability to capture both spatial and temporal dependencies and patterns hidden in the multivariate time series. The latest LLM-based models have shown SOTA performance in major benchmark datasets. They are particularly interested in the zero-shot and few-shot advantages.

%% file: problem.tex
\subsection{Problem definition}
We define the network as a line digraph $L(g) = (N, V)$ where $N$ represents the arcs that connect the nodes in the original directed network graph. Network time series (NTS) are defined as multivariate time series (MTS) data $\mathbb{R}^{T \times N \times C}$ where $T$ represents the time steps of multiple time series, each of which describes $|C|$ performance metrics on a network link $n \in N$. When we are only concerned with a single link metric, such as the traffic volume on a link, we use $n^T$ to represent a single variable TS in the data. The prediction problem is defined as the prediction of the metric values ${n^{T_f}, n \in N}$ of the next $T_f$ future time steps (horizon).

We further define a {\it dependency graph} $G(N,E)$, where the weight function $w(e)$ can be defined to model the dependence relationship between two univariate time series data in $N$.    

\subsection{Time Series Decomposition and Temporal Models}
\begin{figure}
  \begin{center}
    \includegraphics[width=0.48\textwidth]{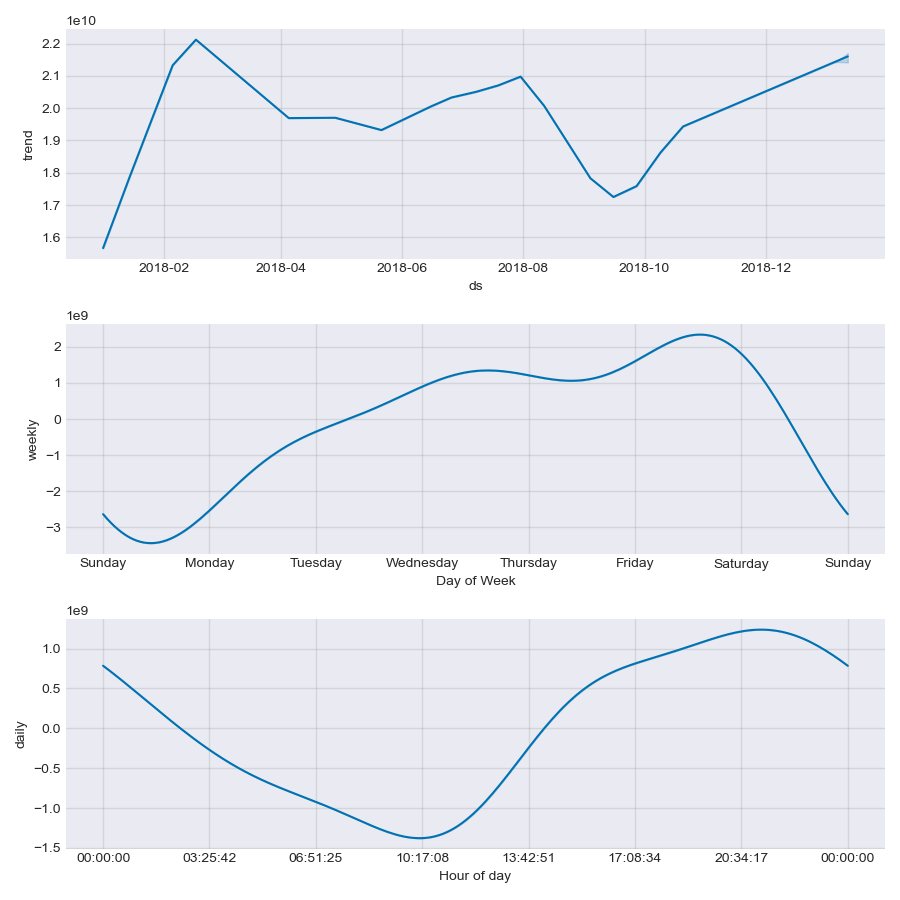}
  \end{center}
  \caption{Time Series Components}
  \label{fig:component}
\end{figure}

Decomposing a time series into trend and seasonality components can provide fundamental insight into the underlying structure of the data. Fig.~\ref{fig:component} shows the decomposition of an individual time on a particular network link. It clearly demonstrates the nonlinear nature of the trend, daily and weekly seasonalities in the three figures, respectively.

As network traffic data possesses complex nonstationary and nonlinear patterns at different time scales, we started our study with the most sophisticated classical SARIMAX univariate time series model enhanced with frequency domain statistics as the exogenous variables, in order to capture the multiple seasonalities. While showing good short-term prediction performance for certain time series, it suffers from deteriorated long-term prediction performance and a large training overhead for a large number of time series individually. On the contrary, the LSTM deep learning model demonstrated overall better performance for both short and long-term predictions. 

Nevertheless, our experiences also indicate that the sheer complexity and scale of the network time series data pose significant challenges to these models in terms of prediction quality and training time. Large network time series exhibit complicated interdependency that may introduce severe overfitting and prolonged training time. Therefore, we focused on more powerful deep learning model architectures with an LSTM model trained as the baseline temporal MTS model.

\subsection{Important Considerations in Model Training and MTS Prediction Performance Evaluation}
Accurate evaluation of multivariate time series prediction models hinges on the choice of performance metrics, especially when dealing with datasets of varying magnitudes such as network traffic. Scale-free metrics like sMAPE (Symmetric Mean Absolute Percentage Error) are preferred because they normalize errors, enabling fair comparisons across time series with different scales. It is robust to zero or near-zero values and provides interpretable, unbiased assessments. In contrast, scale-dependent metrics such as MAE, MSE, and RMSE can disproportionately penalize errors in high-volume series, while MAPE may produce unstable results due to division by small values. Specifically, MTS prediction performance on different time series may vary widely. Our study includes both performance metric selection and distribution evaluation, which have been largely ignored by existing studies. 

Effective training of prediction models requires thoughtful selection of loss functions, optimizers, and preprocessing techniques. The Huber loss function is particularly suitable for real-world data with noise and outliers, offering a balance between the sensitivity of MSE and the robustness of MAE. Optimizers like Adam are commonly used for their adaptability and efficiency. Feature scaling, such as using a Standard Scaler, is essential to ensure stable optimization and prevent features with large numeric ranges from dominating the learning process. Importantly, scalers must be fit only on training data to avoid data leakage and preserve model integrity. To prevent overfitting, regularization techniques, dropout layers, and validation strategies should be employed, especially when dealing with high-dimensional multivariate inputs.

\begin{figure}
  \begin{center}
    \includegraphics[width=0.49\textwidth]{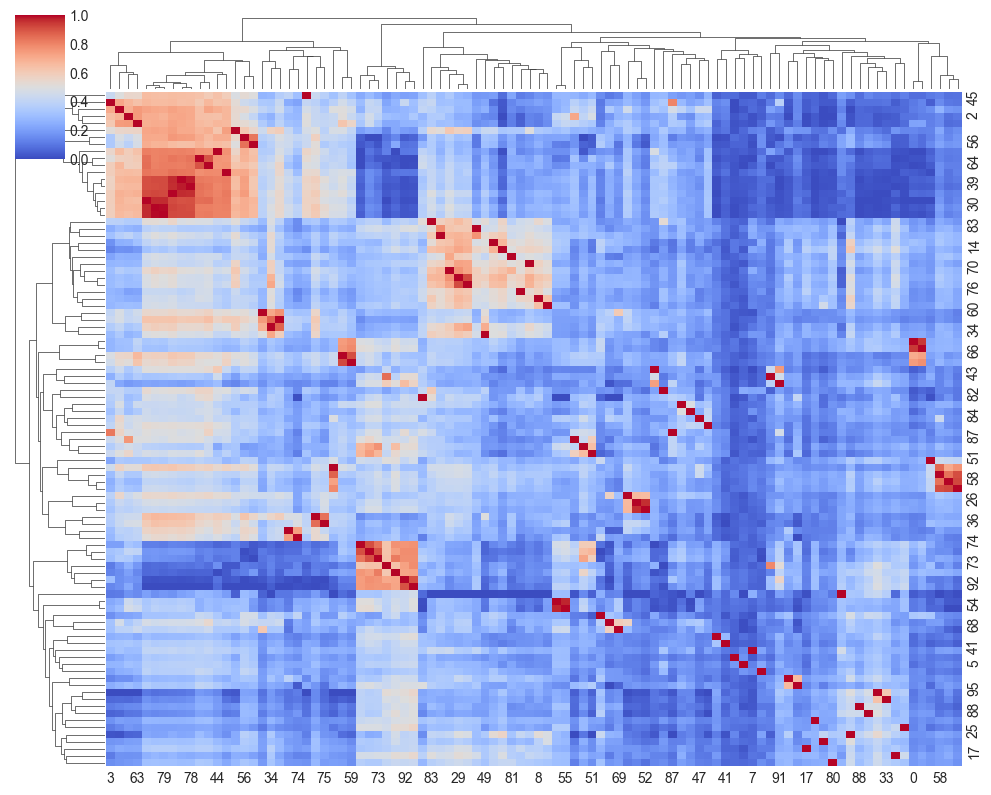}
  \end{center}
  \caption{Network MultiVariate Time Series Cross Correlation}
  \label{fig:dependency}
\end{figure}

Additionally, understanding and leveraging cross-correlation between time series variables can significantly enhance model performance, as it allows the model to capture interdependencies and shared temporal patterns. Fig.~\ref{fig:dependency} depicts the Spearman cross-correlation map with a hierarchical clustering structure on our evaluation network MTS dataset. While there exist strongly correlated clusters among some time series, many of them are not correlated. Unfortunately, this phenomenon of diverse cross-correlation has been largely ignored in existing work. For example, it is easy to see that the time series of many edge links in a network are not likely to be correlated to each other.  

Both spuriously high or misleadingly low correlation estimates can negatively impact model stability and performance. The presence of significant cross-correlation, particularly if redundant or non-linear, often necessitates a strategic reduction in the number of time series used for training and inference. To accurately assess these relationships in real-world data—which often exhibits non-normal distributions, outliers, and non-linear patterns—the Spearman correlation model offers a distinct advantage over the traditional Pearson model. Because it measures monotonic relationships using data ranks, Spearman is more robust to these complexities and provides a reliable basis for identifying true redundancy. Because we are working with time series data, it is natural then to also consider more time-series aware clustering techniques. Techniques like Dynamic Time Warping (DTW) and Shape-Based Clustering (SBC) comes to mind. There would exist tradeoff between the computational complexity and prediction performance gains. As we'll show in the evaluation section, focusing on unique information and mitigating multicollinearity led to more accurate forecasting models. 

The prediction horizon—short-term versus long-term—may also significantly influence model architecture and data requirements. Long-term prediction necessitates models capable of capturing broader seasonal patterns and complex temporal dependencies. We emphasize the difference between the {\it iterative} and {\it direct} long-term prediction. The former uses a one-step-ahead model repeatedly, while the latter is trained to directly output the whole predicted sequence ${n^{T_f}}$, which is more challenging but more useful. Recent advancements in Large Language Models (LLMs) open up new possibilities for treating time series prediction as sequence-to-sequence tasks.

%% file: solution.tex
In this section, we present the core models of the three deep learning architectures for network MTS forecasting. Every architecture starts with the same multivariate input layer and ends with the same output layer over variable horizons.


\subsection{LSTM MTS Model}
\label{subsec:lstm}
We employed a multivariate time series prediction model based on a single-layer Long Short-Term Memory (LSTM) network. The architecture comprises an LSTM layer, followed by a Dropout layer to mitigate overfitting, and a Dense output layer for final predictions. This design enables the model to effectively capture temporal dependencies and complex patterns across multiple input features. All hyperparameters, including sequence length, number of LSTM units, dropout rate, learning rate, batch size, and loss function, were systematically considered, and the optimal model configuration was identified through grid or random search, with selection based on validation performance metrics. The dataset was preprocessed to generate input sequences and corresponding targets, with scaling applied as specified in the configuration. 



\subsection{Network-Temporal Graph Attention Network Model}

\begin{figure}
	\centering
	\includegraphics[width=0.50\textwidth]{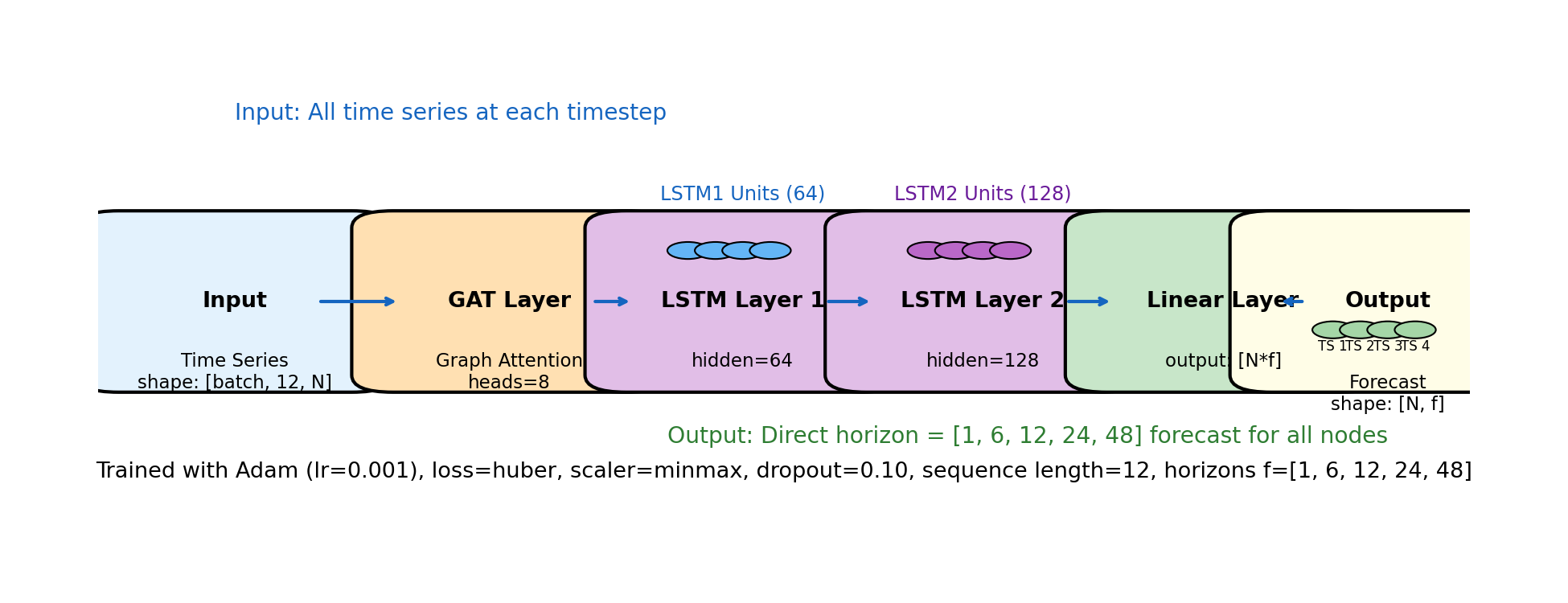}
 \vspace{-0.4in}
	\caption{ST-GAT Model for MTS}
	\label{fig:stgat}
\end{figure}

Fig.~\ref{fig:stgat} shows the spatio-temporal neural network tailored for prediction on graph-structured time series data with optimal hyperparameters. The core spatial component is the Graph Attention (GAT) layer, which uses multi-head attention (typically 8 heads) to dynamically weigh the influence of neighboring nodes, allowing the model to capture both local and global graph dependencies. The GAT layer can be customized with different adjacency modes and hop counts, enabling the model to aggregate information from direct neighbors or more distant nodes, which is crucial for capturing complex network interactions. Following the GAT, two sequential LSTM layers (hidden=64 and hidden=128) model temporal dependencies, and a final linear layer produces multi-horizon predictions.

The number of hops in connecting the nodes in the GAT layer is a critical design choice: increasing hops allows the model to aggregate information from a wider neighborhood, improving performance on networks with long-range dependencies, but may introduce noise if the graph is sparse or poorly connected. We thereafter call this a network-temporal graph attention network (NT-GAT) MTS model. 

\subsection{Multimodal Large Language Model (LLM)}

\begin{figure}
	\centering
	\includegraphics[width=0.50\textwidth]{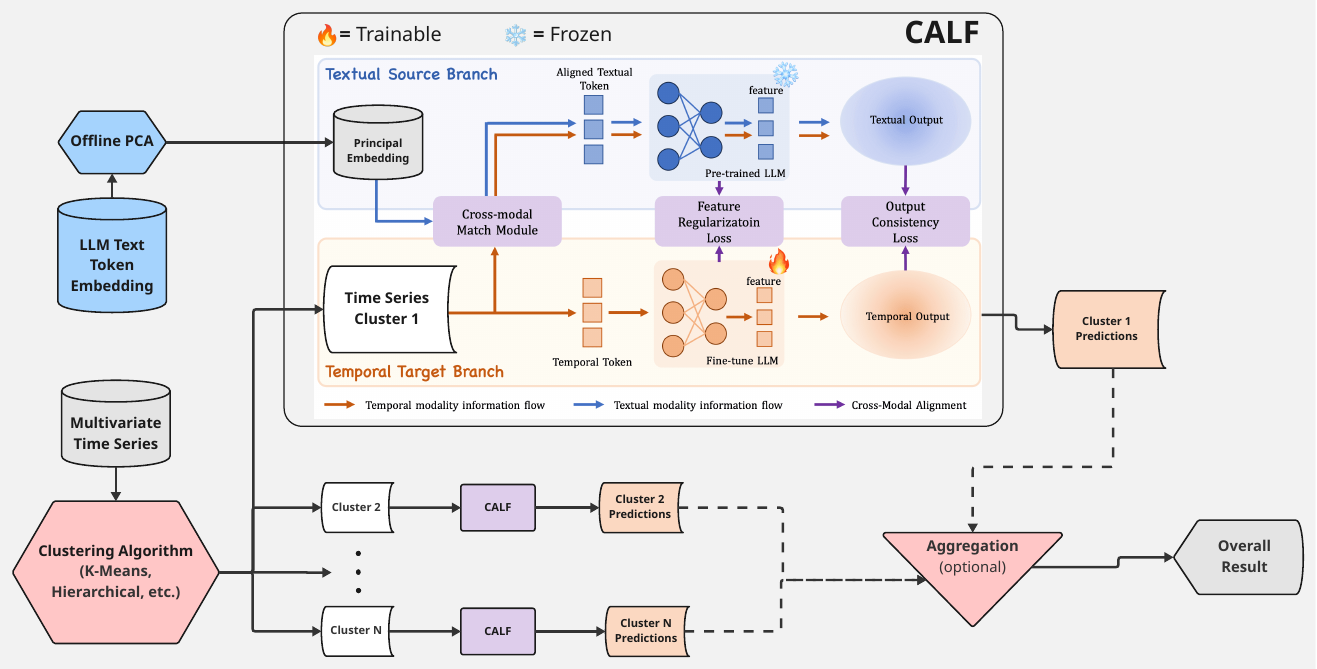}
	\caption{Cluster-CALF: LLMs for Network-Temporal Time Series Prediction via Cross-modal Fine-tuning and Cross-Correlation Clustering}
	\label{fig:calf}
\end{figure}

The CALF (Cross-ModAl LLM Fine-Tuning) model is a novel framework for applying LLMs to MTS prediction with cross-modal fine-tuning capabilities~\cite{liu2025calf}. CALF addresses a key limitation in prior forays into LLM-based prediction methods, that being the distribution discrepancy between the tokens of temporal input data and the textual data on which LLMs were originally trained. Effectively, the framework trains a time-series forecasting branch by forcing it to stay aligned with the LLM’s text embedding space, hidden features, and outputs, so the LLM’s pretrained knowledge transfers more effectively to time series.

As shown within Fig.~\ref{fig:calf}, the CALF framework bridges this gap of aligning the temporal target branch with a textual source branch by using a cross-modal match module to achieve the multi-level alignment. The LLM modified and used in CALF's analysis is OpenAI's GPT2. The choice of using such a model builds on previous work in LLM-based time series analysis, particularly in the work of TimeLLM~\cite{jin2024time} and GPT4TS~\cite{zhou2023one}. A PCA based principle word embedding extraction is performed on the LLM's native word embeddings. A cross-attention mechanism then aligns the projected time-series tokens with these principal word embeddings to generate aligned text tokens for the textual branch. It uses feature regularization loss to align the intermediate hidden states of the two branches, ensuring feature consistency during fine-tuning. Output consistency loss is used to minimize the modality gap in the final output representations. CALF is optimized using parameter-efficient fine-tuning (LoRA) on the temporal branch to enhance training efficiency and preserve the LLM's pre-trained knowledge. The result is a model framework which outperforms both prior LLM and Transformer based forecasting techniques on nearly all of the standard test datasets of ETTH, Weather, ECL, Traffic, and M4~\cite{liu2025calf}.  

\subsection{Enhanced Multimomdal LLM Framework}

Our primary contribution, which we term Cluster-CALF, enhances this architecture by introducing a clustering preprocessing step designed to manage high-dimensional multivariate time series. This addition is motivated by a key challenge in real-world MTS datasets that has been largely ignored: the presence of diverse cross-correlation, where some time series are strongly correlated while many are not. To address this, the enhanced framework, as shown in Fig.~\ref{fig:calf}, first applies an unsupervised clustering analysis to the input features. We select the Spearman cross-correlation as the affinity metric to construct this overture layer. Our experiments validate that it outperforms other clustering methods, such as the Pearson model, due to its rank-based mechanism that is more robust to non-linear patterns, non-normal distributions, and outliers common in network traffic data, as discussed in Section~\ref{sec:problem}. By partitioning the large set of time series into more cohesive clusters based on these robustly-identified relationships, our approach reduces input complexity and mitigates multicollinearity, which could otherwise interfere with CALF's ability to effectively forecast on the multivariate time series. Each cluster is then fed individually into the CALF framework, enabling the model to more efficiently learn the intra-cluster temporal patterns.



%% file: evaluation.tex
We use an hourly traffic volume dataset over a year from an Internet backbone service provider's network, which features close to one hundred bi-directional links. For long-term horizons, we use the {\it direct} prediction methodology and employ a training-validation-testing pipeline with data splitting. Both local and high-performance computing (SLURM) environments with GPUs were utilized to perform hyperparameter sweeps and final training runs, ensuring consistency in the training pipeline. All training logs and results were systematically recorded for subsequent analysis. For prediction performance evaluation, we use sMAPE (Symmetric Mean Absolute Percentage Error) as our main metric, as it normalizes errors and mitigates bias from varying time series scales—an important consideration for heterogeneous network data. Our analysis mainly focused on accuracy and so an in-depth efficiency analysis was not conducted, thus constituting future work in the analysis of our framework. 

An analysis of traditional statistical methods or a comparison against existing LLM or transformer-based deep learning models is not included in this paper. On the front of traditional statistical methods, a comprehensive analysis of statistical methods was conducted, but none were able to outperform LSTM. Due to space limitations, the data is not included here. We also studied existing transformer-based models. However, the base LLM architecture (CALF) which we leverage was able to outperform all existing LLM-based and transformer-based models on nearly all test datasets~\cite{liu2025calf}. Thus, we maintain it suffices to show that our new framework which we propose is able to outperform the base CALF model. 

\subsection{LSTM MTS Model}

\begin{figure}
  \begin{center}
    \includegraphics[width=0.49\textwidth]{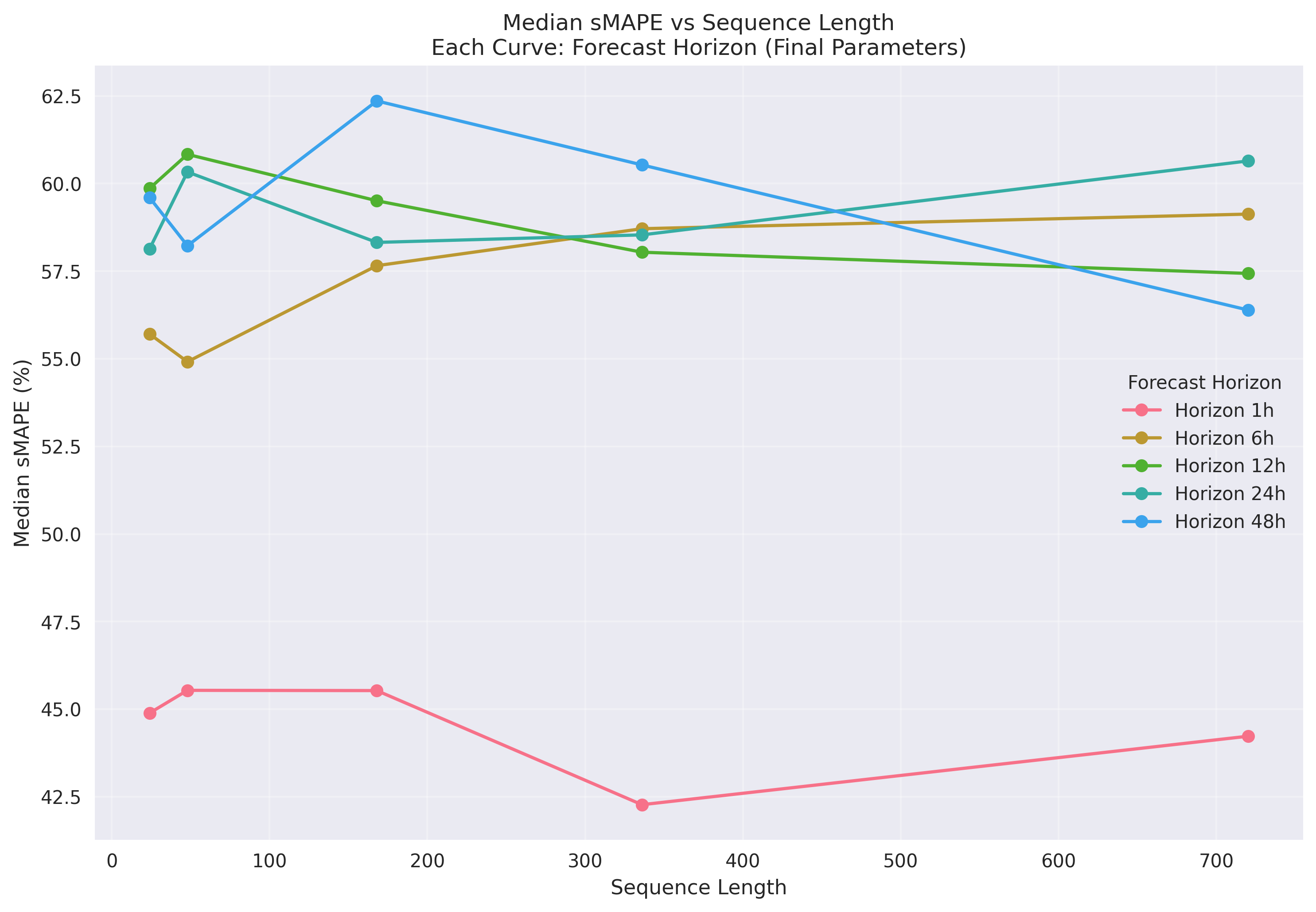}
  \end{center}
  \caption{LSTM prediction Performance vs. Prediction Horizons}
  \label{fig:hori:lstm}
\end{figure}

\begin{figure}
  \begin{center}
    \includegraphics[width=0.49\textwidth]{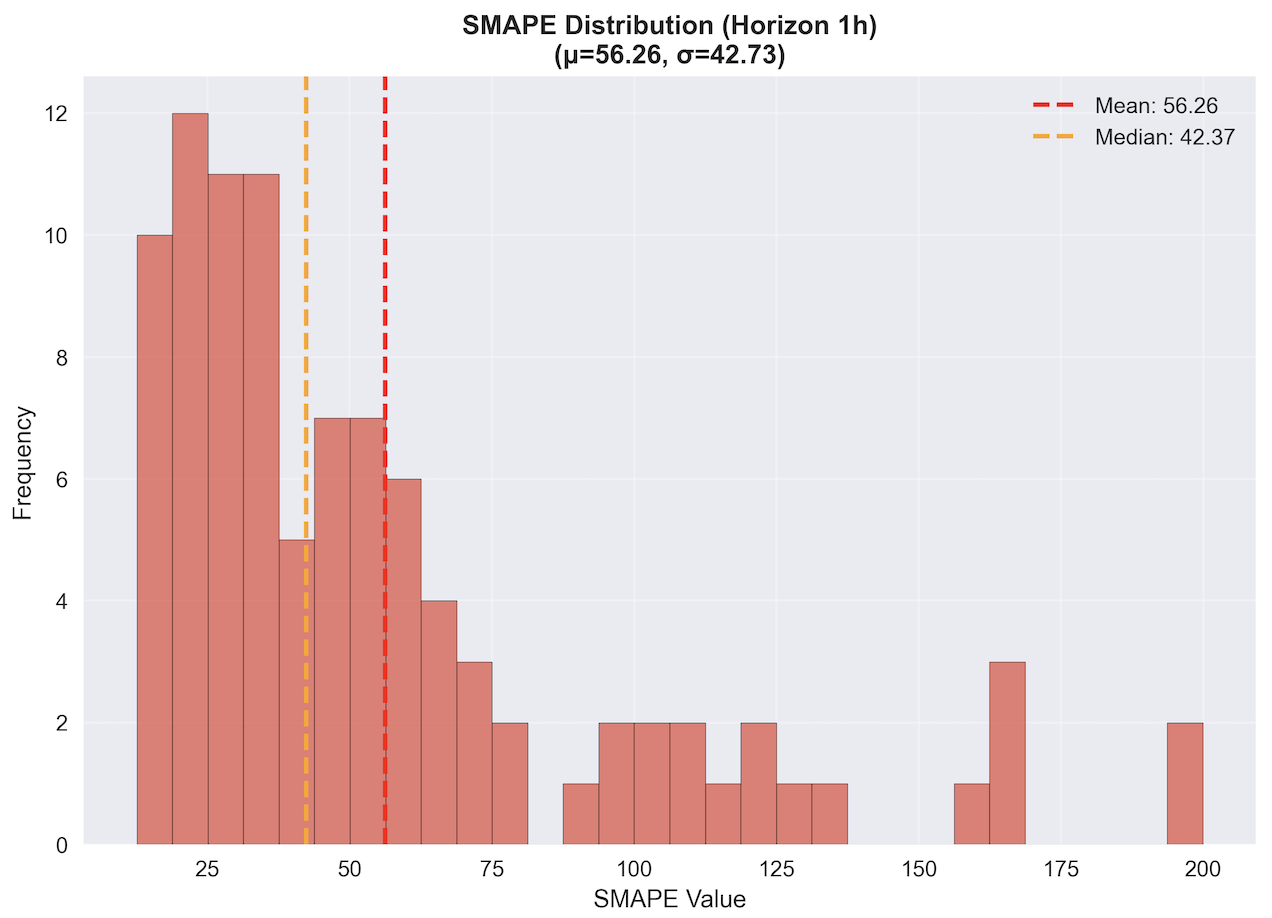}
  \end{center}
  \caption{LSTM Prediction Performance Distribution}
  \label{fig:dist:lstm}
\end{figure}

Fig.~\ref{fig:hori:lstm} and ~\ref{fig:dist:lstm} show the overall prediction performance of the LSTM model with the best hyperparameters after an extensive grid-search, as shown in Fig~\ref{fig:hori:lstm}.

Rather than averaging results over multiple horizons, we analyzed the model’s prediction performance across all time series columns over multiple prediction horizons with varying input sequence lengths. While the best horizon, determined to be one hour with a mean sMAPE of approximately $56.26\%$ with sequence length of 336, longer term predictions deteriorate noticeably, though their performance differences do not show clear patterns and the impact of sequence length is not significant. With Fig.~\ref{fig:dist:lstm}, we want to draw attention to the widely distributed performance among the time series. The gap between the median and mean indicates outliers. Note that analysis of traditional statistical methods was conducted, but none are able to outperform the LSTM MTS model, which we use as our baseline. 

\subsection{NT-GAT MTS Model}

\begin{figure}
  \begin{center}
\includegraphics[width=0.49\textwidth]{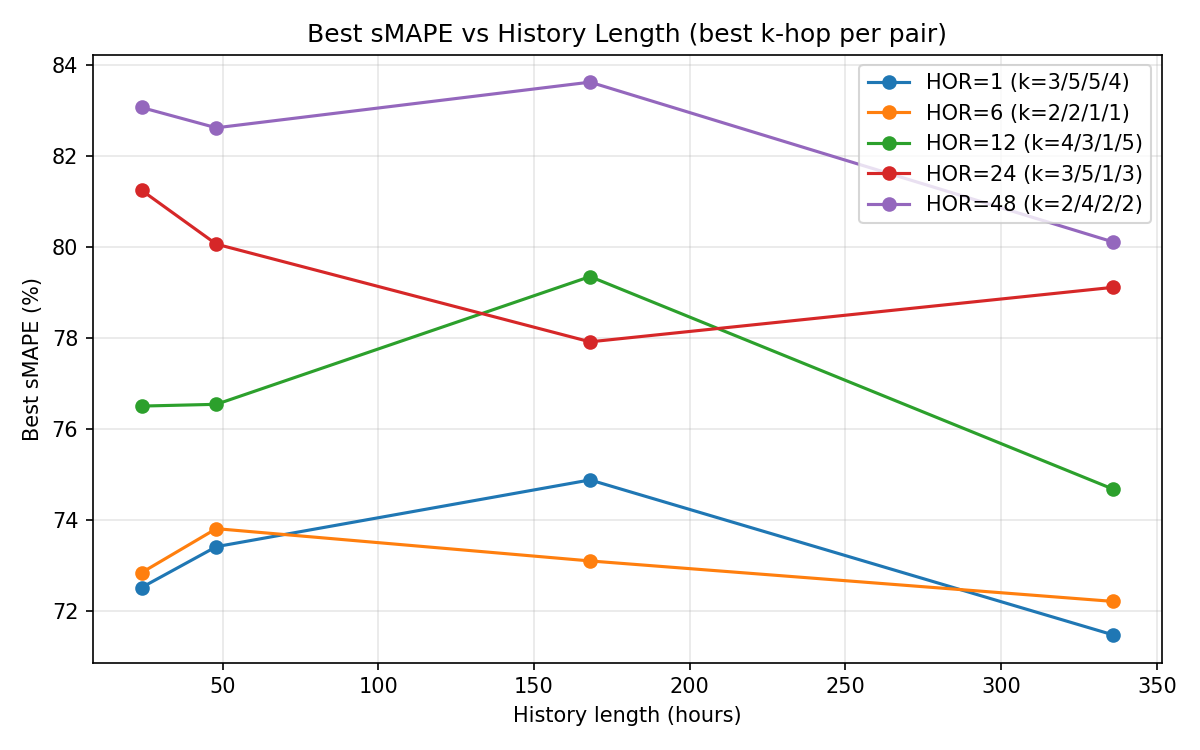}
  \end{center}
  \caption{ST-GAT Prediction Performance vs. Prediction Horizons}
  \label{fig:hori:stgat}
\end{figure}

\begin{figure}
  \begin{center}
    \includegraphics[width=0.49\textwidth]{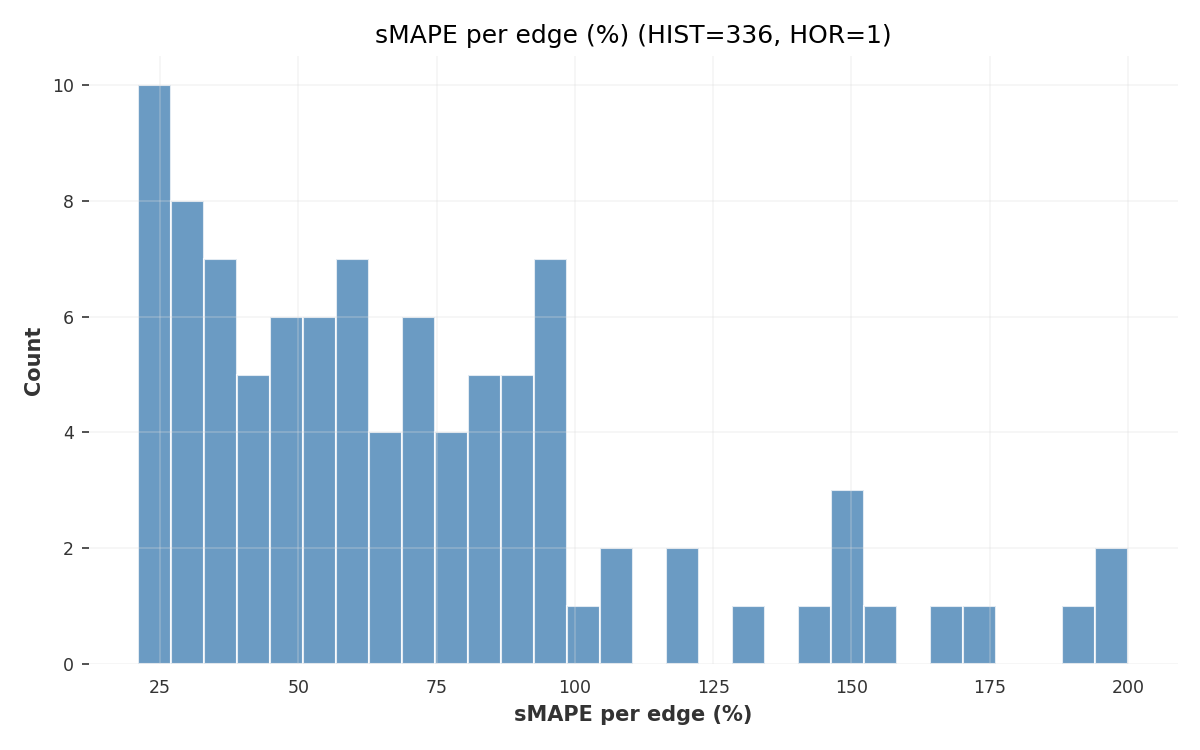}
  \end{center}
  \caption{ST-GAT Prediction Performance Distribution}
  \label{fig:dist:stgat}
\end{figure}

We focus on the impacts of sequence length and the graph connectivity hops on the performance of our NT-GAT model across multiple horizons. Error distributions (sMAPE, MAE) are consistently lower for nodes with higher connectivity, confirming the benefit of graph-based aggregation. Ablation studies reveal that Huber loss improves stability and generalization compared to other loss functions. The choice of hop count in the GAT layer directly impacts the model’s ability to capture network effects, with very short hop ($2-5$)aggregation yielding the best trade-off between local and global information for this network dataset. Fig.~\ref{fig:hori:stgat} and ~\ref{fig:dist:stgat} show the overall prediction performance of the ST-GAT MTS Model. Noticeably, the average performance of this much more complicated model is worse than that of the simpler LSTM model, probably an indication of overfitting. On the other hand, the performance difference between different horizons is much smaller, and the distribution among time series is concentrated more towards the smaller value, a positive indication of the effect of the attention mechanism.

\subsection{Cluster-CALF MTS Model}

\begin{figure}
  \begin{center}
\includegraphics[width=0.49\textwidth]{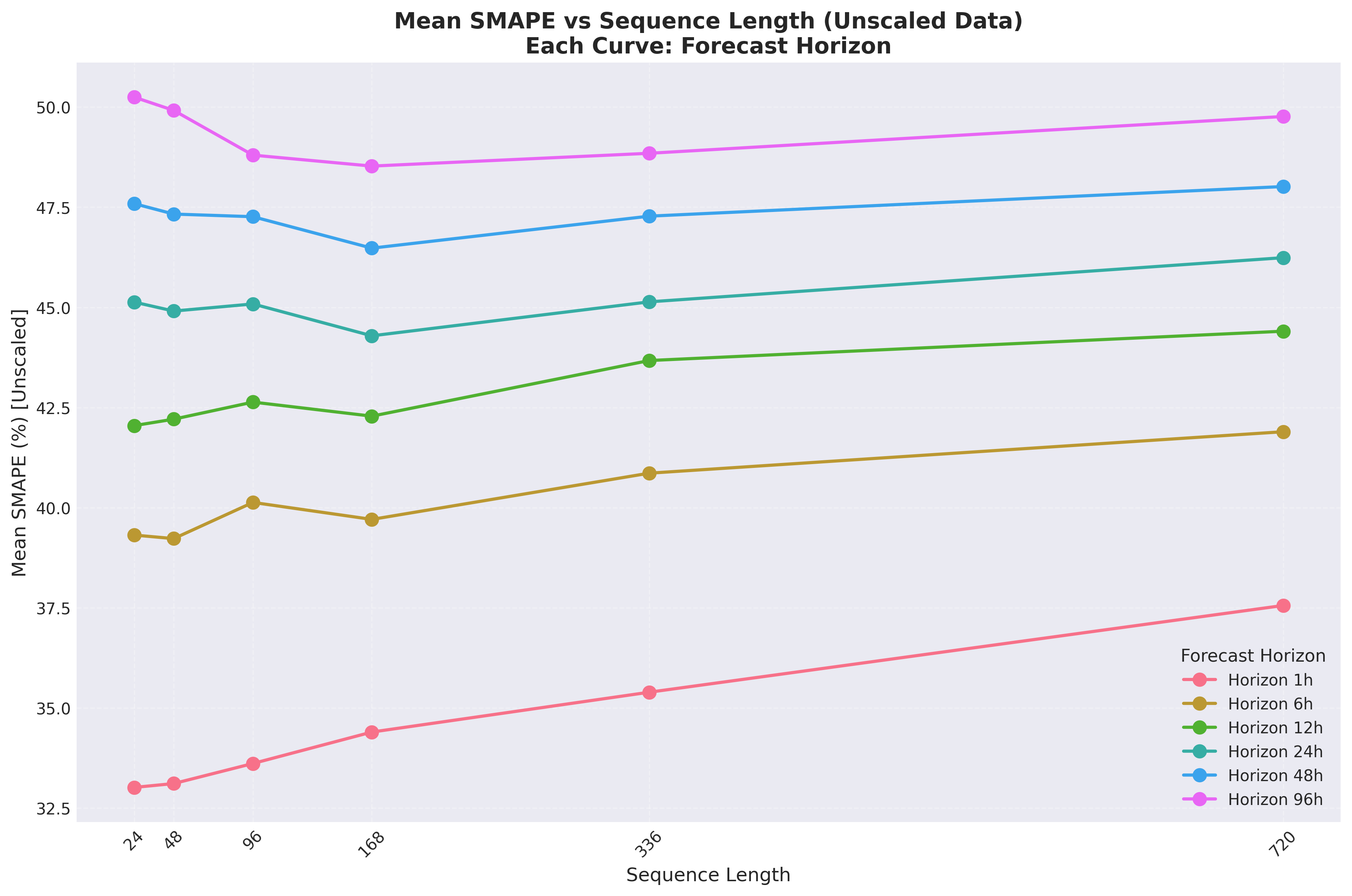}
  \end{center}
  \caption{CALF Prediction Performance vs. Prediction Horizons}
  \label{fig:hori:llm}
\end{figure}

We first consider the base CALF model with no clustering preprocessing step to determine the best hyperparameters, as we have done with LSTM and ST-GAT. Fig.~\ref{fig:hori:llm} shows the overall prediction performance of the CALF model, with a grid search over the same prediction horizons and sequence length pairs as LSTM. Similar to LSTM, we observe that a prediction of a horizon of 1 yields the best results across all sequence lengths. The CALF model outperforms both LSTM and ST-GAT at their best mean sMAPE points, yielding a  $41.31\%$ decrease in best mean sMAPE compared to LSTM. Upon further analysis, we also observe that the distribution of sMAPEs across columns for the LLM MTS is also tighter, with a $29\%$ decrease in standard deviation. Thus, we confirm the effectiveness of the CLAF model over LSTM and ST-GAT for our prediction task.

\begin{figure}
  \begin{center}
    \includegraphics[width=0.49\textwidth]{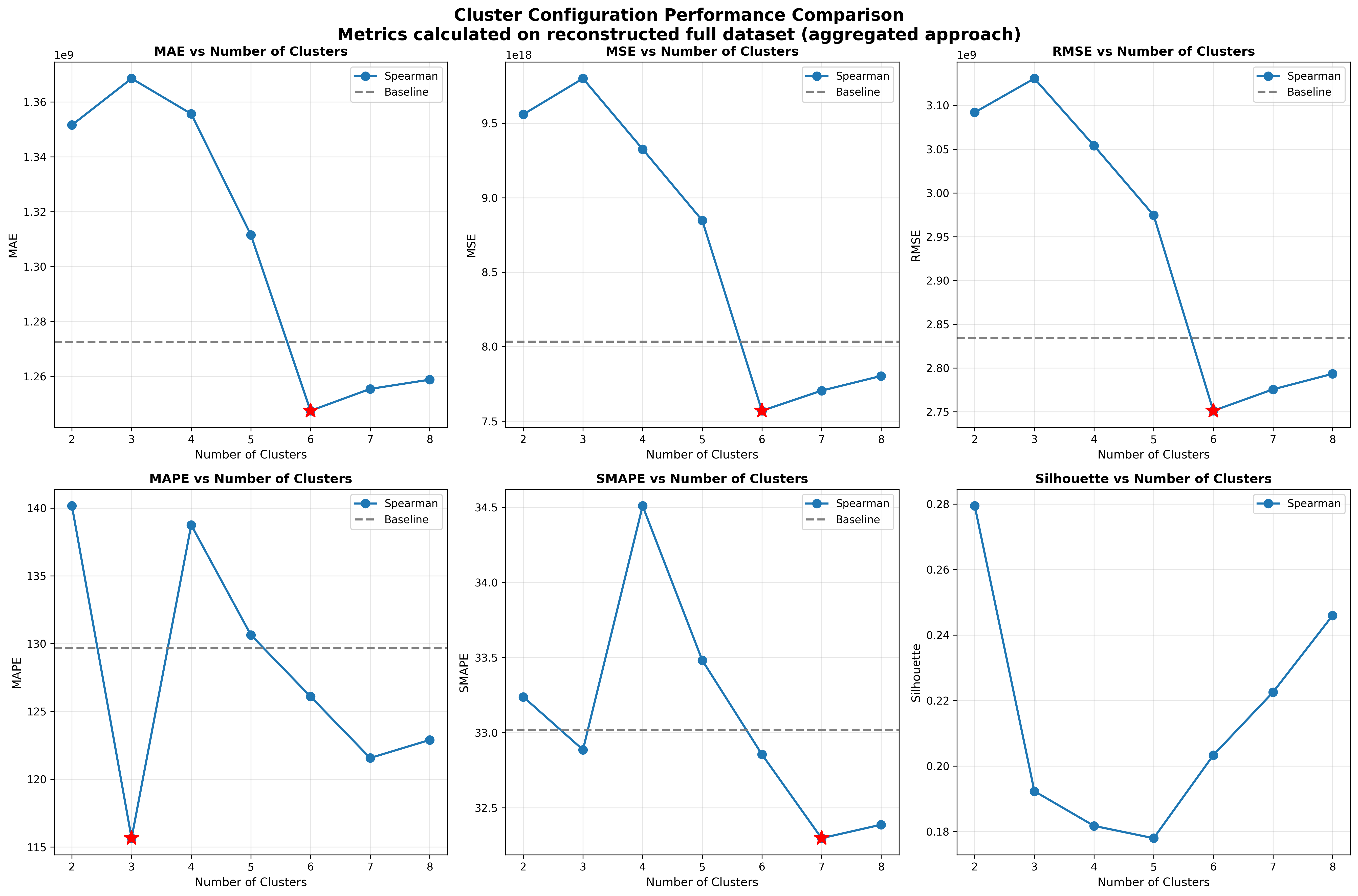}
  \end{center}
  \caption{Cluster-CALF Performance vs. Number of Clusters}
  \label{fig:comp:llm}
\end{figure}

\begin{figure}
  \begin{center}
    \includegraphics[width=0.49\textwidth]{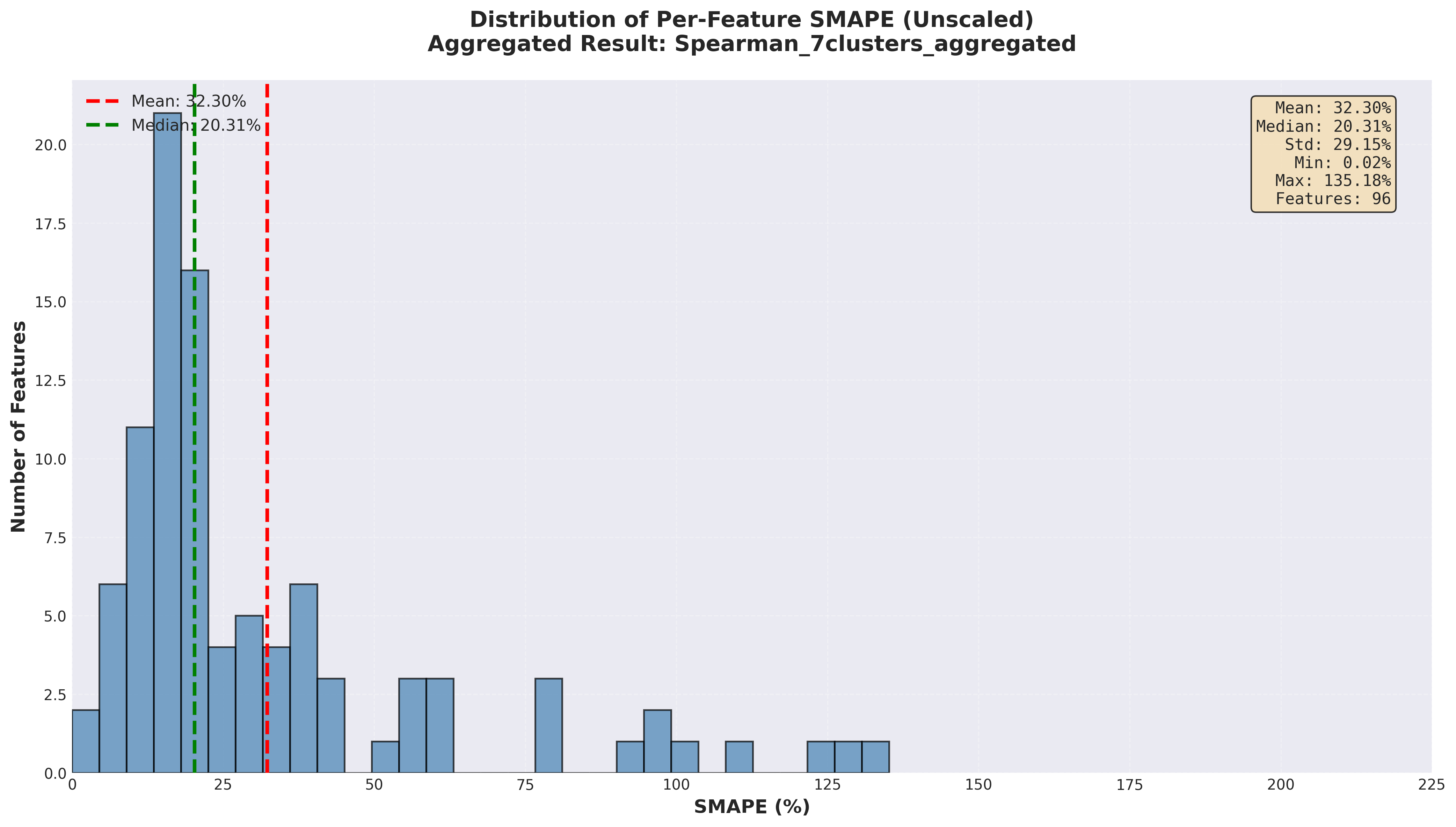}
  \end{center}
  \caption{Cluster-CALF Prediction Performance Distribution}
  \label{fig:dist:llm}
\end{figure}

\begin{figure}
  \begin{center}
    \includegraphics[width=0.49\textwidth]{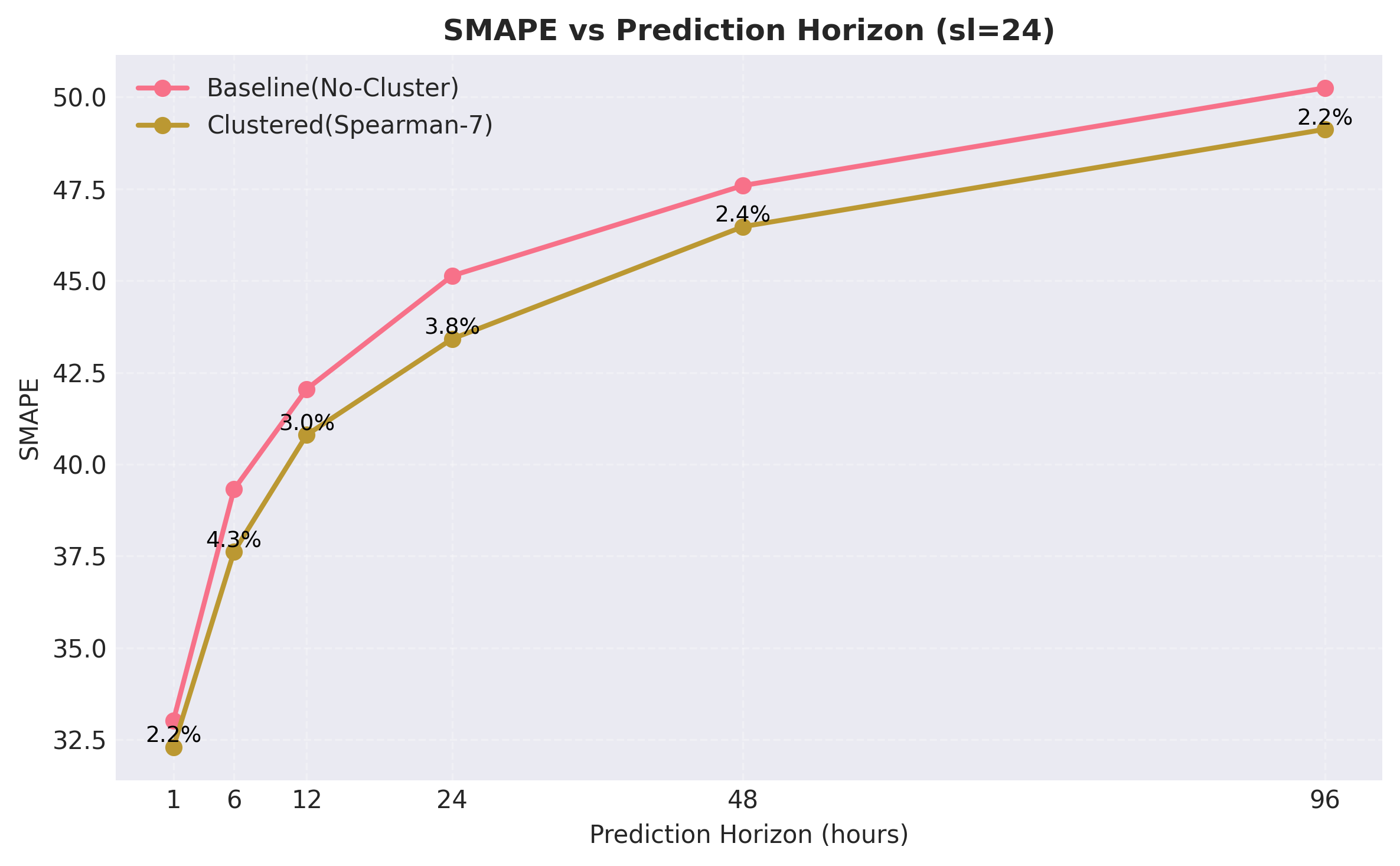}
  \end{center}
  \caption{Cluster-CALF and CALF: Prediction Horizon vs. sMAPE over Spearman Clustering (7 clusters) with Fixed Sequence Length}
  \label{fig:comp:predcluster}
\end{figure}

From there, we perform Spearman clustering analysis over the best-performing configuration (prediction horizon=1, sequence length=24). Fig.~\ref{fig:comp:llm} shows the comparison of cluster configuration performances. The baseline performance (from Fig.~\ref{fig:hori:llm}) is indicated by a dotted line for comparison. We can see that for all error metrics, clusters with sizes of 6, 7, and 8 outperform the baseline. In particular, we observe that for the best performing cluster over sMAPE, 7 clusters as shown in Fig.~\ref{fig:dist:llm}, we see both a $2.18\%$ decrease in mean sMAPE along with a $3.41\%$ decrease in the standard deviation of the distribution. When performing analysis of 7 Spearman clusters over a range of prediction horizons, as shown in Fig.~\ref{fig:comp:predcluster}, the results show a consistent decrease in sMAPE across all tested prediction horizons when compared to the single forecast baseline. More improvements are made in the mid-range, with a peak of $4.3\%$ mean sMAPE decrease at a horizon of 6. 

These results validate the proposed Cluster-CALF framework as a two-stage success. First, the baseline CALF model itself is established as a highly effective forecaster, supplanting previous methods with a $41.31\%$ sMAPE reduction over LSTM. Second, our analysis demonstrates that this already-strong model can be both further and consistently improved with the addition of our clustering preprocessing step.

%% file: future.tex
Prediction is a foundational function of AI-driven network intelligence for next-generation infrastructure. The goal of our research presented in this paper is to identify the best multivariate time series prediction model that can effectively learn the intricate network topological interdependency and generalize the complex temporal patterns in network traffic data. Our primary contribution lies in the creation and evaluation of an enhanced multi-modal LLM framework using unsupervised clustering overture to sharpen the focus of the LLM architecture, displaying superior overall performance compared to other deep learning models on real-world network traffic data. With the important insights gained from the detailed analysis in correlation variability and distribution discrepancies among time series over short- and long-term prediction horizons, our future work will be focused on enhanced prediction performance and model generalization. In particular, additional clustering or alternative preprocessing techniques are a major part of our contribution which can be further explored. Network traffic data in particular presents opportunities to exploit the graph representation of the data, something which our framework has yet to explore. There is also work to be done in the realm of efficiency analysis.
Additional clustering methodologies, including time-series aware Dynamic Time Warping~\cite{wu2020fastdtw} and unsupervised learning models, are being evaluated. While the primary benefit of cluster-based input partitioning is the observed accuracy improvement, the efficiency implications of running multiple smaller forecasting jobs versus a single larger job remains to be studied. Future work should empirically characterize the computational tradeoffs of this approach, and investigate whether cluster number and composition can be chosen to jointly optimize for both accuracy and efficiency.

%% file: network.bib
@article{ali2024resource,
  title={A resource-aware multi-graph neural network for urban traffic flow prediction in multi-access edge computing systems},
  author={Ali, Ahmad and Ullah, Inam and Shabaz, Mohammad and Sharafian, Amin and Khan, Muhammad Attique and Bai, Xiaoshan and Qiu, Li},
  journal={IEEE Transactions on Consumer Electronics},
  year={2024},
  publisher={IEEE}
}

@article{tanaka2021monitoring,
	author = {Tanaka, Takafumi and Inui, Tetsuro and Kawai, Shingo and Kuwabara, Seiki and Nishizawa, Hideki},
	journal = {Journal of Optical Communications and Networking},
	number = {10},
	publisher = {Optica Publishing Group},
	title = {Monitoring and diagnostic technologies using deep neural networks for predictive optical network maintenance},
	volume = {13},
	year = {2021}}

@article{shaghaghi2021proactive,
	author = {Shaghaghi, Amirhossein and Zakeri, Abolfazl and Mokari, Nader and Javan, Mohammad Reza and Behdadfar, Mohammad and Jorswieck, Eduard A},
	journal = {IEEE Transactions on Network and Service Management},
	number = {1},
	publisher = {IEEE},
	title = {Proactive and AoI-aware failure recovery for stateful NFV-enabled zero-touch 6G networks: Model-free DRL approach},
	volume = {19},
	year = {2021}}

@inproceedings{plan:drl:zhu2021,
	author = {Zhu, Hang and Gupta, Varun and Ahuja, Satyajeet Singh and Tian, Yuandong and Zhang, Ying and Jin, Xin},
	booktitle = {Proceedings of the 2021 ACM SIGCOMM 2021 Conference},
	title = {Network planning with deep reinforcement learning},
	year = {2021}}

@inproceedings{singh2022traffic,
	author = {Singh, Rachee and Bj{\o}rner, Nikolaj and Krishnaswamy, Umesh},
	booktitle = {Proceedings of the Symposium on SDN Research},
	title = {Traffic engineering: from ISP to cloud wide area networks},
	year = {2022}}


%% file: tmgpt.bib
@article{wu2020fastdtw,
  title={FastDTW is approximate and generally slower than the algorithm it approximates},
  author={Wu, Renjie and Keogh, Eamonn J},
  journal={IEEE Transactions on Knowledge and Data Engineering},
  volume={34},
  number={8},
  pages={3779--3785},
  year={2020},
  publisher={IEEE}
}

@inproceedings{liu2025calf,
  title={Calf: Aligning llms for time series forecasting via cross-modal fine-tuning},
  author={Liu, Peiyuan and Guo, Hang and Dai, Tao and Li, Naiqi and Bao, Jigang and Ren, Xudong and Jiang, Yong and Xia, Shu-Tao},
  booktitle={Proceedings of the AAAI Conference on Artificial Intelligence},
  volume={39},
  number={18},
  year={2025}
}

@ARTICLE{st_gan,
  author={Zhang, Chenhan and Yu, James J. Q. and Liu, Yi},
  journal={IEEE Access}, 
  title={Spatial-Temporal Graph Attention Networks: A Deep Learning Approach for Traffic Forecasting}, 
  year={2019},
  volume={7},
  number={},
  doi={10.1109/ACCESS.2019.2953888}}

@inproceedings{lai2018modeling,
	author = {Lai, Guokun and Chang, Wei-Cheng and Yang, Yiming and Liu, Hanxiao},
	booktitle = {The 41st international ACM SIGIR conference on research \& development in information retrieval},
	title = {Modeling long-and short-term temporal patterns with deep neural networks},
	year = {2018}}

@article{mohammed2021predicting,
	author = {Mohammed, Bashir and Krishnaswamy, Nandini and Kiran, Mariam and Wu, Keshang},
	journal = {International Journal of Big Data Intelligence},
	number = {1},
	publisher = {Inderscience Publishers (IEL)},
	title = {Predicting WAN traffic volumes using Fourier and multivariate SARIMA approach},
	volume = {8},
	year = {2021}}

@inproceedings{jin2024time,
	author = {Jin, Ming and Wang, Shiyu and Ma, Lintao and Chu, Zhixuan and Zhang, James and Shi, Xiaoming and Chen, Pin-Yu and Liang, Yuxuan and Li, Yuan-fang and Pan, Shirui and others},
	booktitle = {International Conference on Learning Representations},
	title = {Time-LLM: Time Series Forecasting by Reprogramming Large Language Models},
	year = {2024}}

@inproceedings{NEURIPS2023_3eb7ca52,
	author = {Gruver, Nate and Finzi, Marc and Qiu, Shikai and Wilson, Andrew G},
	booktitle = {Advances in Neural Information Processing Systems},
	editor = {A. Oh and T. Naumann and A. Globerson and K. Saenko and M. Hardt and S. Levine},
	publisher = {Curran Associates, Inc.},
	title = {Large Language Models Are Zero-Shot Time Series Forecasters},
	volume = {36},
	year = {2023}}

@article{garza2023timegpt,
	author = {Garza, Azul and Mergenthaler-Canseco, Max},
	journal = {arXiv preprint arXiv:2310.03589},
	title = {TimeGPT-1},
	year = {2023}}

@inproceedings{ilbert2023breaking,
	author = {Ilbert, Romain and Hoang, Thai V and Zhang, Zonghua and Palpanas, Themis},
	booktitle = {Proceedings of the 2023 Workshop on Recent Advances in Resilient and Trustworthy ML Systems in Autonomous Networks},
	title = {Breaking Boundaries: Balancing Performance and Robustness in Deep Wireless Traffic Forecasting},
	year = {2023}}

@article{zhou2023one,
	author = {Zhou, Tian and Niu, Peisong and Sun, Liang and Jin, Rong and others},
	journal = {Advances in neural information processing systems},
	title = {One fits all: Power general time series analysis by pretrained lm},
	volume = {36},
	year = {2023}}

@article{ts:forecast:pitfalls:2023,
	author = {Hewamalage, Hansika and Ackermann, Klaus and Bergmeir, Christoph},
	journal = {Data Mining and Knowledge Discovery},
	number = {2},
	publisher = {Springer},
	title = {Forecast evaluation for data scientists: common pitfalls and best practices},
	volume = {37},
	year = {2023}}
